\documentclass[11pt]{article}
\usepackage{CJK}
\usepackage[utf8]{inputenc}
\usepackage{coling2018}
\usepackage{times}
\usepackage{url}
\usepackage{latexsym}
\usepackage{amsmath}
\usepackage{graphicx}
\usepackage{subcaption}
\usepackage{natbib}
\usepackage{pgfplots}
\usepackage{pgfplotstable}
\pgfplotsset{width=6.0cm,compat=1.5}
\usepackage{color,xcolor,framed}

\newcommand*{\affaddr}[1]{#1} 
\newcommand*{\affmark}[1][*]{\textsuperscript{#1}}
\newcommand*{\email}[1]{\texttt{#1}}

\title{Deconvolution-Based Global Decoding for Neural Machine Translation}

\author{Anonymous authors}
\author{Junyang Lin\affmark[1,2], Xu Sun\affmark[2], Xuancheng Ren\affmark[2], Shuming Ma\affmark[2], Jinsong Su\affmark[3], Qi Su\affmark[1]\\
\affaddr{\affmark[1]School of Foreign Languages, Peking University}\\
\affaddr{\affmark[2]MOE Key Lab of Computational Linguistics, School of EECS, Peking University}\\
\affaddr{\affmark[3]School of Software, Xiamen University}\\
\email{\{linjunyang, xusun, renxc, shumingma, sukia\}@pku.edu.cn}\\
\email{jssu@xmu.edu.cn}\\
}

\date{}

\begin{document}
\maketitle
\begin{CJK}{UTF8}{gbsn}
\begin{abstract}
  A great proportion of sequence-to-sequence (Seq2Seq) models for Neural Machine Translation (NMT) adopt Recurrent Neural Network (RNN) to generate translation word by word following a sequential order. As the studies of linguistics have proved that language is not linear word sequence but sequence of complex structure, translation at each step should be conditioned on the whole target-side context. To tackle the problem, we propose a new NMT model that decodes the sequence with the guidance of its structural prediction of the context of the target sequence. Our model generates translation based on the structural prediction of the target-side context so that the translation can be freed from the bind of sequential order. Experimental results demonstrate that our model is more competitive compared with the state-of-the-art methods, and the analysis reflects that our model is also robust to translating sentences of different lengths and it also reduces repetition with the instruction from the target-side context for decoding.
\end{abstract}

\section{Introduction}
\label{intro}

%
%
\blfootnote{
    \hspace{-0.65cm}  
    This work is licensed under a Creative Commons
    Attribution 4.0 International License.
    License details:
    \url{http://creativecommons.org/licenses/by/4.0/}
}

Deep learning has achieved tremendous success in machine translation, outperforming the traditional linguistic-rule-based and statistical methods. In recent studies of Neural Machine Translation (NMT), most models are based on the sequence-to-sequence (Seq2Seq) model based on the encoder-decoder framework \citep{Kalchbrenner,seq2seq,ChoEA2014} with the attention mechanism \citep{attention,stanfordattention}. While traditional linguistic-rule-based and statistical methods of machine translation require much work of feature engineering, NMT can be trained in the end-to-end fashion. Besides, the attention mechanism can model the alignment relationship between the source text and translation \citep{attention,stanfordattention}, and some recent improved versions of attention have proved successful in this task \citep{coverage, micover,interactive,multichannel,googleattention}.

However, the decoding pattern of the recent Seq2Seq models is inconsistent with the linguistic analysis. As the conventional decoder translates words in a sequential order, the current generation is highly dependent on the previous generation and it is short of the knowledge about future generation. \citet{nida} pointed out that translation goes through a process of analysis, transfer and reconstruction, involving the deep syntactic and semantic structure of the source and target languages. Language generation involves complex syntactic analysis and semantic integration, instead of a step-by step word generation \citep{Frazier}. Moreover, from the perspective of semantics and pragmatics, the syntactic analysis of utterance can be guided by the global lexical-semantic and discourse information \citep{Altmann, Trueswellsemantic, Trueswellverb, Tyler}. In brief, the process of translation is in need of the global information from the target-side context, but the decoding pattern of the conventional Seq2Seq model in NMT does not meet the requirement.

Recent researches in NMT have taken this issue into consideration by the implementation of bidirectional decoding. Some methods of bidirectional decoding \citep{Liu2016Agreement,Cong2017Towards} rerank the candidate translations with the scores from the bidirectional decoding. However, these bidirectional decoding methods cannot provide effective complementary information due to the limited search space of beam search.


In this article, we extend the conventional attention-based Seq2Seq model by introducing the deconvolution-based decoder, which is a Convolutional Neural Network (CNN) to perform deconvolution. Recently, deconvolution has been applied to the studies of natural language \citep{deconvolution,deconvseq}, which can be regarded as the transposition of the convolution \citep{FCN, deconv_ss}. \citet{deconvolution} applied this method to natural language by modeling sentences with a convolution-deconvolution autoencoder. The study of \citet{deconvolution} showed that the deconvolution solves the problems by reconstructing a representation of high quality irrespective to the order or length. It can be found that deconvolution owns the potential to provide global information for guidance of decoding. Therefore, we follow this idea and propose a new model with deconvolution for NMT.

To be specific, the conventional RNN encoder encodes the source sentences to new representations and sends the final state to the decoder, and the conventional RNN decoder decodes it to the target sentences with the attention to the encoder outputs. In our model, our designed deconvolution-based decoder decodes the final state of the encoder to a matrix representing the global information of the target-side contexts. Each column of the matrix is learned to be close to the word embedding of the target words. The conventional RNN decoder can attend to the columns for the information of the target-side context to perform global decoding in the translation.

Our contributions in this study are illustrated in the following:
\begin{itemize}
\item We propose a new model for NMT, which contains a deconvolution-based decoder to provide global information of the target-side contexts to the RNN decoder, so that the model is able to perform global decoding\footnote{The code is released in \url{https://github.com/lancopku/DeconvDec}}.
\item Experimental results demonstrate that our model outperforms the baseline models in both the Chinese-to-English translation and the English-to-Vietnamese translation, outperforming the Seq2Seq model in the BLEU score evaluation with the advantages of BLEU score 2.82 and 1.54 respectively.
\item The analysis shows that our model that performs global decoding is more capable of reducing repetition and more robust to the translation of sentences of different lengths, and the case study reflects that it is able to capture the syntactic structure for the translation and has a better reflection of the semantic meaning of the source text.
\end{itemize}

\section{Model}

\begin{figure*}[t]
\centering
\includegraphics[width=0.8\linewidth]{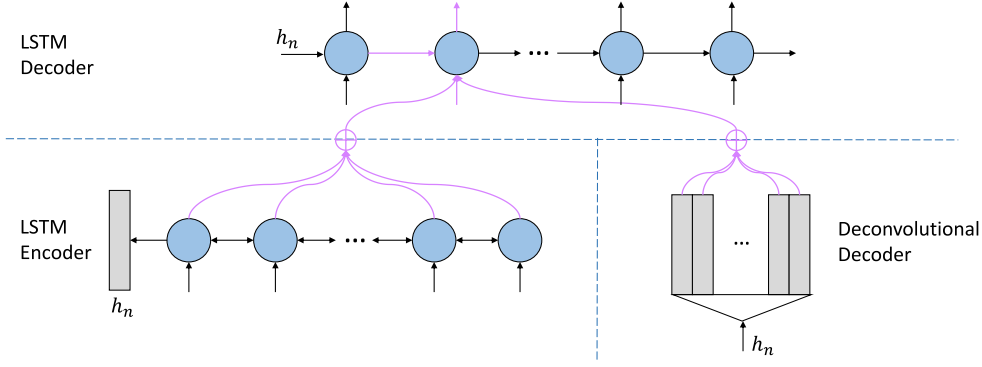}
\caption{Model architecture. There are three components in the proposed model, i.e., the LSTM encoder, the deconvolution-based decoder, and the conventional LSTM decoder. The encoder distills the input sentence into a state $h_n$, which is then used in the deconvolution-based decoder to obtain the global information of the target-side contexts. Based on the target-side contexts and the input-side contexts, the conventional LSTM decoder generates the output from the state $h_n$.}
\end{figure*}

In the following, we introduce the details of our model, including the encoder, the deconvolution-based decoder and the conventional RNN-based decoder. The functions of each decoder are illustrated below to show how they collaborate to improve the quality of the translation.

\subsection{Encoder}
In our model, the encoder reads the embeddings of the input text sequence ${x = \{x_{1}, ..., x_{n}\}}$ and encodes a sequence of encoder outputs ${h = \{h_{1}, ..., h_{n}\}}$. The final hidden state ${h_{n}}$ is sent to the decoder as the initial state for it to decode a sequence of output text. The encoder outputs provide the information of the source-side contexts to our RNN-Based decoder through the attention mechanism.

The encoder in our model is a bidirectional LSTM \citep{LSTM},
which reads the input in two directions to generate two sequences of hidden states $\overrightarrow{{h}} \!=\! \{\overrightarrow{{h_{1}}}, \overrightarrow{{h_{2}}}, \overrightarrow{{h_{3}}}, ..., \overrightarrow{{h_{n}}}\}$ and $\overleftarrow{{h}} \!=\! \{\overleftarrow{{h_1}}, \overleftarrow{{h_2}}, \overleftarrow{{h_3}}, ... ,\overleftarrow{{h_n}}\}$, where:
\begin{align}
\overrightarrow{{h_{i}}} = {LSTM}({x_{i}}, \overrightarrow{{h_{i-1}}},{C_{i-1}}) \label{eq7}\\
\overleftarrow{{h_{i}}} = {LSTM}({x_{i}}, \overleftarrow{{h_{i-1}}}, {C_{i-1}}) \label{eq8}
\end{align}
The encoder outputs corresponding to each time step are concatenated as ${h_i}\!=\![\overrightarrow{{h_i}};\overleftarrow{{h_i}}] \label{eq9}$.

\subsection{Deconvolution-Based Decoder}

In our model, there are two decoders, which perform different tasks for the whole decoding process. While the RNN-based decoder is similar to the conventional decoder, which decodes the output text sequence in a sequential order and attends to the annotations of the encoder via the attention mechanism, our proposed deconvolution-based decoder does not decode the text but provide global information of the target-side contexts to the RNN-based decoder so that it can decode structurally instead of sequentially. To be specific, the deconvolution-based decoder learns to generate the word embedding matrix of the target text sequence.

In order to provide global information of the target-side contexts to the RNN decoder, we implement a multilayer CNN as the deconvolution-based decoder to perform deconvolution. With deconvolution, it is available for a vector or a small matrix to be transformed to a large matrix. In our model, the deconvolution is implemented on the final states in both directions from the encoder. As words in our model are represented with word embedding vectors, sentences can be formed as word embedding matrices. In our model, the deconvolution-based decoder is designed to learn word embedding matrices of the target sequences with the representation matrix from the encoder. As the conjugate operation of convolution, deconvolution expands the dimension of the input representation to a matrix of our designed size. There are $L$ layers in the deconvolution, each of which has $f_{l}$ filters of kernel size $k_{l}$. The $i^{th}$ filter $W_{l}^{i} \in {R}^{{k} \times {dim}}$ ($dim$ refers to the size of the input representation vector) with stride $s_{l}^{i}$ and padding $p_{l}^{i}$ performs deconvolution on the input representation matrix $I \in {R}^{{m} \times {dim}}$ ($m \times {dim}$ refers to the size of the input representation matrix), the final hidden state of the encoder. The computation of convolution is illustrated as below:
\begin{align}
c^{i}_{l} = g(W_{l}^{i} * X + b)
\end{align}
where $X$ refers to the convolved matrix and $g$ refers to non-linear activation function, which is ReLU \citep{ReLU} following \citet{deconvolution}, and deconvolution is its transposed operation. With the input $I$, our objective of the deconvolution operation is to generate a word embedding matrix $E \in {R}^{{T} \times {dim}}$ where $T$ refers to the sentence length designed for the output text sequence, which is a hyper-parameter. At the $l^{th}$ layer, deconvolution generates a matrix $E_{l} \in {R}^{{T_{l}} \times {dim}}$ where $T_{l} = T_{l-1} \times s_{l} + k_{l} - 2 \times p_{l}$. With the control of stride and kernel size, the height of the matrix can be assigned, and with the control of the number of filters, the width of the matrix can also be assigned, which are the length of the output sequence and the dimension of the word embedding respectively.

The deconvolution-based decoder can generate meaningful representation with information different from that in the conventional RNN decoder. The conventional RNN decoder generates sequence in a way similar to Markov Decision Process, which is highly dependent on the previous generation and follows a strict sequential order, so it contains high sequential dependency. On the contrary, the deconvolution-based decoder generates a word embedding matrix depending on the representation from the encoder without considering the order, which does not have the problem of long-term dependency. Moreover, although it is not capable as the RNN encoder to generate coherent text, it reveals the information of the text from a global perspective, including syntactic and semantic features.

\begin{figure*}[t]
\centering
\subcaptionbox{Deconvolution-based decoder.}{\includegraphics[width=0.5\linewidth]{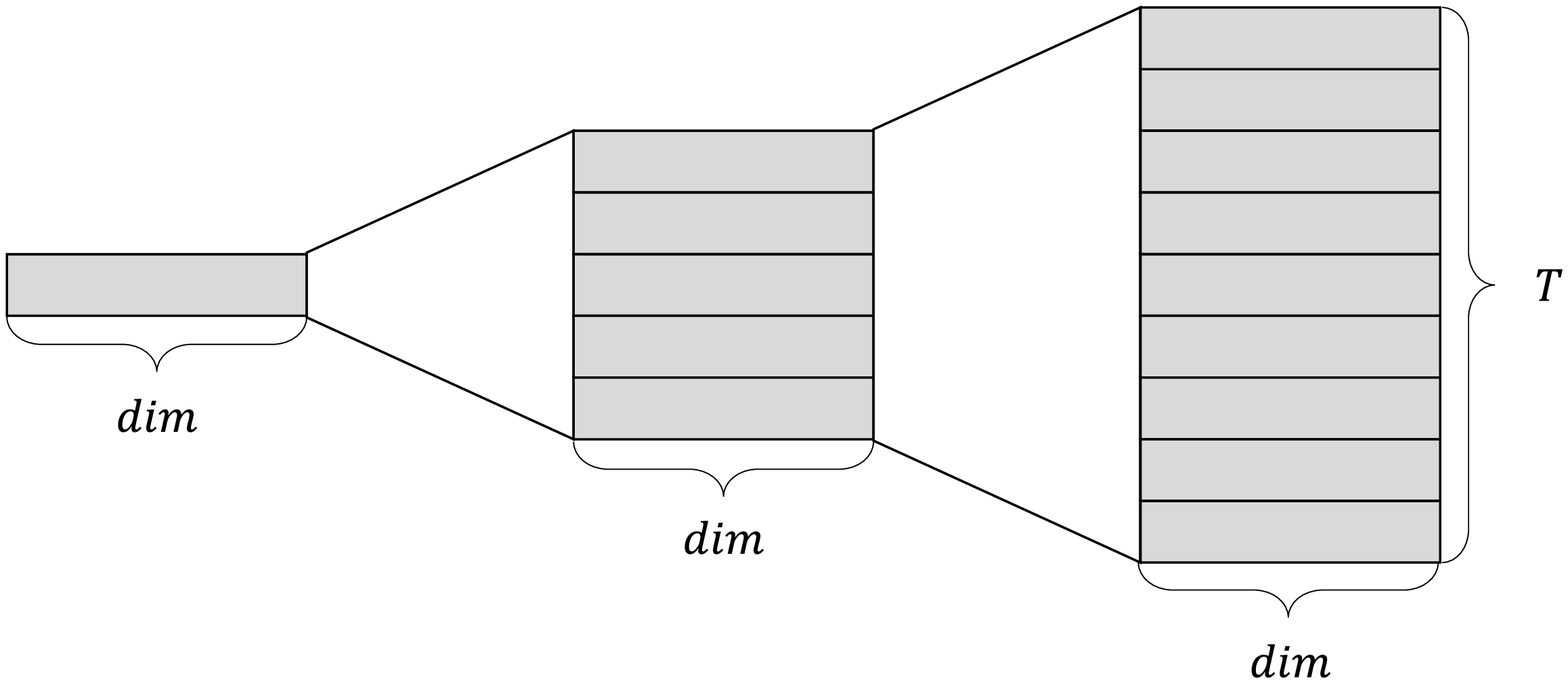}} \hspace{1cm}
\subcaptionbox{Deconvolution.}{\includegraphics[width=0.35\linewidth]{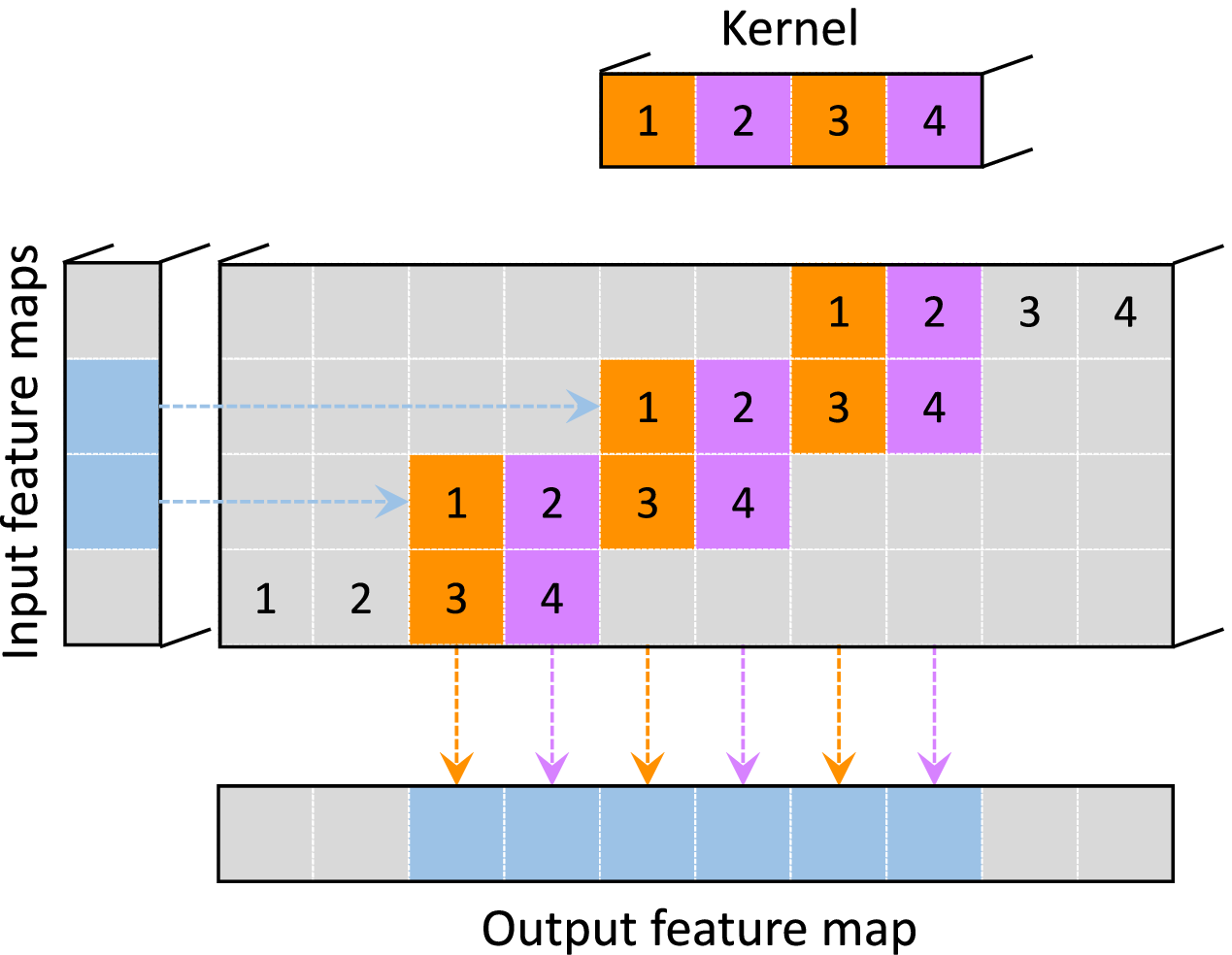}}
\caption{Deconvolution-based decoder. On the right shows an example of a 1d deconvolution on a input of size 2 with a kernel of size 4, a padding of $1$ and, a stride of $2$. The depth means the dimension of the channel, which is $dim$ in our case.}
\end{figure*}

\subsection{RNN-Based Decoder}
Different from the deconvolution-based decoder, the RNN-based decoder is responsible for decoding the representation ${h_{n}}$ to generate the translation ${y = \{y_{1}, ..., y_{m}\}}$. With the final encoder state as the initial state, the decoder is initialized to decode in sequential order, until it generates the token representing the end of sentence. During decoding, the attention mechanism is applied for the decoder to extract the information from the source-side contexts, which are the annotations of the encoder, as well as the information from the target-side contexts, which are the outputs of the deconvolution-based decoder.

For the RNN-based decoder, we implement a unidirectional LSTM. The output of the RNN-based decoder at each time step is sent into a feed-forward neural network to be projected into the space of the target vocabulary ${Y} \in {R}^{\mid{Y}\mid \times {dim}}$ for the prediction of the translated word. At each time step, the decoder generates a word ${y_{t}}$ by sampling from a conditional probability distribution of the target vocabulary ${P_{vocab}}$, where: 
\begin{align}
{P_{vocab}} &= {softmax}({W_{o}v_{t}}) \label{predict}\\
{v_{t}} &= {g(s_{t}, c_{t}^{}, \tilde{c}_{t})} \\
{s_{t}} &= {LSTM}({y_{t-1}, s_{t-1}}, {C_{t-1}}) \label{eq12}
\end{align}
where ${g(\cdot)}$ refers to non-linear activation function, and $c_{t}$ and $\tilde{c}_{t}$ are the outputs of the attention mechanism, which are illustrated in the following.

The attention mechanism in our model is the global attention mechanism \citep{stanfordattention}. Different from the conventional attention mechanism, which only computes the attention scores on the source-side contexts, the attention mechanism in our model consists of two parts. The first one is similar to the conventional one, attending to the source-side contexts from the encoder, but the second one is original, which attends to the target-side contexts, which is the word embedding matrix generated by the deconvolution-based decoder. By attending to the encoder annotations, the model computes the attention ${\alpha_{t,i}}$ of the RNN-based decoder output ${s_{t}}$ on the annotations of the encoder $h_{i}$ and generates the context vector ${c_{t}}$. Similarly, by attending to the outputs of the deconvolution-based decoder, the RNN-based decoder computes the attentions of $s_{t}$ on each column $E_{i}$ of its matrix $E$ and generates the context vector $\tilde{c}_{t}$:
\begin{align}
c_{t} &= {\sum^{n}_{i=1} \alpha_{t,i}h_{i}} \label{eq13}\\
\tilde{c}_{t} &= {\sum^{n}_{i=1} \tilde{\alpha}_{t,i}E_{i}}
\end{align}
where $\alpha_{t,i}$ and $\tilde{\alpha}_{t,i}$ are defined as below (as they are computed in the same way, they are both represented by $\alpha_{t,i}$ and the annotations are represented by $x_{i}$):
\begin{align}
\alpha_{t,i} &= \frac{{exp}({e_{t,i}})}{{\sum_{j=1}^{n}}{exp}({e_{t,j}})} \label{eq14}\\
{e_{t,i}} &= {s_{t-1}^{\top}}{W_{a}x_{i}} \label{eq15}
\end{align}

\subsection{Training}

The training for the Seq2Seq model is usually based on maximum likelihood estimation. Given the parameters $\theta$ and source text $x$, the model generates a sequence $\tilde{y}$. The learning process is to minimize the negative log-likelihood between the generated text $\tilde{y}$ and reference $y$, which in our context is the sequence in target language for machine translation:
\begin{align}
\mathcal{L} &= -\frac{1}{{N}}{\sum_{i=1}^{N}}{\sum_{t=1}^{T}}{\log P(y_{t}^{(i)}}|{\tilde{y}_{<t}^{(i)},x^{(i)}, \theta)} \label{eq22}
\end{align}
where the loss function is equivalent to maximizing the conditional probability of sequence $y$ given parameters $\theta$ and source sequence $x$.

However, as there are two decoders in our model, the loss function should also be designed for the deconvolution-based decoder. In our model, we compute the smooth L1 loss between the generated matrix of the deconvolution-based decoder $E$ and the word embedding matrix $\tilde{E}$ (which both contain $M$ elements), which is more robust to outliers \citep{fastRCNN}, as well as the cross-entropy loss between the prediction of the deconvolution-based decoder $\hat{y}$ and reference $y$ given the parameters of the encoder and the deconvolution-based decoder $\theta^{'}$. Therefore, the generated matrix $E$ can be closer to the word embedding matrix $\tilde{E}$, and it contains information beneficial to the prediction of the target words. Moreover, for the cross entropy loss of the deconvolution-based decoder, we apply the method of \citet{wean} as it increases no parameter for the prediction by computing the cosine similarity between the output and the word embeddings. To sum up, the loss function is defined as below:
\begin{align}
\mathcal{L} &= -\frac{1}{{N}}{\sum_{i=1}^{N}}({\sum_{t=1}^{T}}{\log P(y_{t}^{(i)}}|{\tilde{y}_{<t}^{(i)},x^{(i)}, \theta)} + {\sum_{m=1}^{M}} smooth_{L1}(E_{m}-\tilde{E}_{m}) + {\sum_{t=1}^{T}}{\log P(y_{t}^{(i)}}|x^{(i)},\theta^{'}))
\end{align}
where smooth L1 loss is defined below:
\begin{align}
smooth_{L1}(x, y) &= \left\{
\begin{array}{lcl}
             0.5 \mid \mid x-y \mid \mid_{2}^{2} &\text{if} \mid \mid x-y \mid \mid < 1 \\
             \mid \mid x - y \mid \mid_{1} - 0.5 &\text{if} \mid \mid x-y \mid \mid \ge 1
\end{array} 
\right.
\end{align}

We have tested L1 loss, L2 loss as well as smooth L1 loss in our experiments and found that smooth L1 loss encourages the model to reach the best performance.

\section{Experiment}

\subsection{Datasets}
We evaluate our proposed model on the NIST translation task for the Chinese-to-English translation and provide the analysis on the same task. Moreover, in order to evaluate the performance of our model on the low-resource translation, we also evaluate our model on the IWLST 2015 \citep{2015iwslt} for the English-to-Vietnamese translation task.

\textbf{Chinese-to-English Translation} For the NIST translation task, we train our model on 1.25M sentence pairs extracted from LDC2002E18, LDC2003E07, LDC2003E14, Hansards portion of LDC2004T07, LDC2004T08 and LDC2005T06, with 27.9M Chinese words and 34.5M English words. Following \citet{memdec}, we validate our model on the dataset for the NIST 2002 translation task and tested our model on that for the NIST 2003, 2004, 2005, 2006 translation tasks. We use the most frequent 30K words for both the Chinese vocabulary and the English vocabulary, which includes around 97.4\% and 99.5\% of the Chinese and English words in the training data. The sentence pairs longer than 50 words are filtered. The evaluation metric is BLEU \citep{bleu}.

\textbf{English-to-Vietnamese Translation} The data is from the translated TED talks, containing around 133K training sentence pairs provided by the IWSLT 2015 Evaluation Campaign \citep{2015iwslt}. We follow the studies of \citet{nplm}, and use the same preprocessing methods as well as the same validation and the test set. The validation set is the TED tst2012 with 1553 sentences and the test set is the TED tst2013 with 1268 sentences. The English vocabulary is 17.7K words and the Vietnamese vocabulary is 7K words. The evaluation metric is also BLEU score.

\subsection{Setting}
We implement the models on PyTorch\footnote{\url{http://pytorch.org}}, and the experiments are conducted on an NVIDIA 1080Ti GPU. Both the size of word embedding and the number of units in the hidden layers are 512, and the batch size is 64. We use Adam optimizer \citep{KingmaBa2014} to train the model with the setting $\beta_{1}=0.9$, $\beta_{2}=0.98$ and $\epsilon=1\times10^{-9}$ following \citet{googleattention}, and we initialize the learning rate to 0.0003. 

Gradient clipping is applied so that the norm of the gradients cannot be larger than a constant, which is 10 in our experiments. Dropout is used with the dropout rate set to 0.3 for the Chinese-to-English translation and 0.4 for the English-to-Vietnamese translation, based on the performance on the validation set. 

Based on the performance on the validation set, we use beam search with a beam width of 10 to generate translation for the evaluation and test, and we normalize the log-likelihood scores by sentence length.

\subsection{Baselines}

For the Chinese-to-English translation, we compare our model with the state-of-the-art NMT systems for the task.
\begin{itemize}
\item \textbf{Moses} An open source phrase-based translation system
with default configurations and a 4-gram language model trained on the training data for the target language;
\item \textbf{RNNsearch} An attention-based Seq2Seq with fine-tuned hyperparameters \citep{attention};
\item \textbf{Coverage} The method extends RNNSearch with a coverage model for the attention mechanism that tackles the problem of over-translation and under-translation \citep{coverage};
\item \textbf{Lattice} The Seq2Seq model with a word-lattice-based RNN encoder that tackles the problem of tokenization in NMT \citep{lattice};
\item \textbf{InterAtten} The Seq2Seq model that records the interactive history of decoding \citep{interactive};
\item \textbf{MemDec} Based on the RNNSearch, it is equipped with external memory that the model reads and writes during decoding \citep{memdec}.
\end{itemize}

For the English-to-Vietnamese translation, we compare our model with the recent NMT models for this task, and we present the results of the baselines reported in their articles.
\begin{itemize}
\item \textbf{RNNsearch-1} The attention-based Seq2Seq model by \citet{luong2015stanford};
\item \textbf{RNNsearch-2} The implementation of the attention-based Seq2Seq by \citet{nplm};
\item \textbf{LabelEmb} Extending RNNSearch with soft target representation \citep{labelemb};
\item \textbf{NPMT} The Neural Phrased-based Machine Translation model by \citet{nplm};
\end{itemize}

\begin{table*}[tb]
\centering
    \begin{tabular}{l|c|c|c|c|c}
    \hline
    Model  & MT-03 & MT-04 & MT-05 & MT-06 & Ave.\\ \hline\hline
    Moses  & 32.43 & 34.14 & 31.47 & 30.81 & 32.21        \\ 
    RNNSearch  & 33.08 & 35.32 & 31.42 & 31.61 & 32.86        \\ 
    Lattice   &  34.32 &  36.50 &  32.40 & 32.77  &  34.00  \\
    Coverage  & 34.49 & 38.34 & 34.91 & 34.25 & 35.49 \\
    InterAtten  & 35.09 & 37.73 & 35.53 & 34.32 & 35.67 \\
    MemDec  & 36.16 & \textbf{39.81} & 35.91 & 35.98 & 36.97 \\
     \hline\hline
     Seq2Seq+Attention & 35.32  & 37.25 & 33.52  & 33.54  & 34.91 \\
    \textbf{+DeconvDec}  & \textbf{38.04}  & 39.75 & \textbf{36.77}  &  \textbf{36.32} & \textbf{37.73} \\
    \hline
    \end{tabular}
    \caption{Results of our model and the baselines (the results are those reported in the referred articles, and the models are trained on the identical training data or larger training data) on the Chinese-to-English translation, tested on the NIST Machine Translation tasks in 2003, 2004, 2005, 2006 by BLEU score evaluation.}
    \label{cnen}
\end{table*}

\begin{table}[tb]
\centering
    \begin{tabular}{l|c}
    \hline
    Model & BLEU  \\ \hline\hline
    RNNSearch-1  &    23.30          \\ 
    RNNSearch-2  &     26.10                 \\ 
    LabelEmb  &     26.80                 \\
    NPMT &   27.69               \\
    \hline\hline
     Seq2Seq+Attention &  26.93 \\
    \textbf{+DeconvDec} &  \textbf{28.47} \\
    \hline
    \end{tabular}
    \caption{Results of our model and the baselines (directly reported in the referred articles) on the English-to-Vietnamese translation, tested on the TED tst2013 with the BLEU score evaluation.}
    \label{envi}
\end{table}

\section{Results and Analysis}
\subsection{Results}
Table \ref{cnen} shows the overall results of the models on the Chinese-to-English translation task. Beside our reimplementation of the attention-based Seq2Seq model, we report the results of the recent NMT models, which are results in their original articles or improved results of the reimplementation. To facilitate fair comparison, we compare with the baselines that are trained on the same training data. The results have shown that for the NIST 2003, 2004, 2005 and 2006 translation tasks, our model with the deconvolution-based decoder outperforms the baselines, and the advantage of BLEU score over the attention-based Seq2Seq model is 2.82 on average compared with our reimplementation of the attention-based Seq2Seq model. From the results mentioned above, it can be inferred that the global information of the target-side contexts retrieved from the deconvolution-based decoder is contributive to the translation. Our analysis and case study in the following can further demonstrate how the deconvolution-based decoder improves the attention-based Seq2Seq model.

Table \ref{envi} presents the results of the models on the English-to-Vietnamese translation. Compared with the attention-based Seq2Seq model, including the implementation with the strongest performance, our model with the deconvolution-based decoder can outperform it with the advantage of BLEU score 1.54. We also display the most recent model NPMT \citep{nplm} trained and tested on the dataset. Compared with NPMT, our model has an advantage of BLEU score 0.78. It can be indicated that for the low-resource translation, the information from the deconvolution-based decoder is important, which brings significant improvement to the conventional attention-based Seq2Seq model.

\subsection{Analysis}
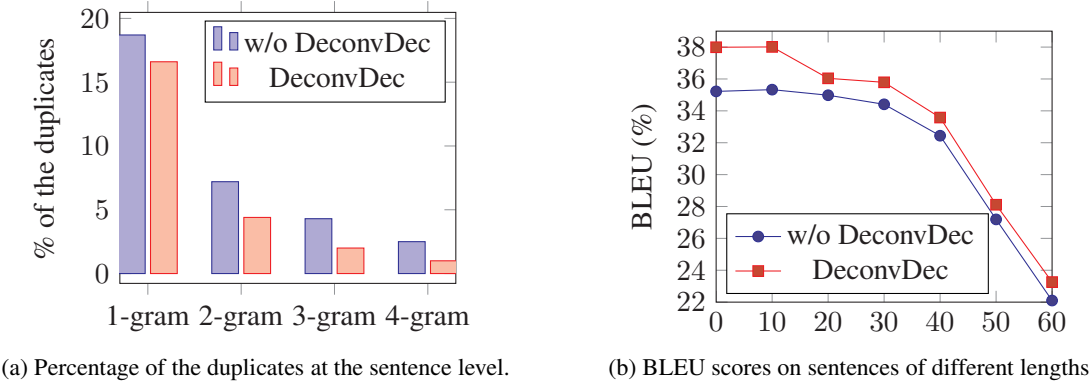
\begin{figure}[tb]
\captionsetup[subfigure]{font=footnotesize}
\centering
\subcaptionbox{Percentage of the duplicates at the sentence level.}[.45\textwidth]{%
\begin{tikzpicture}
\selectcolormodel{cmyk}
\begin{axis}[legend pos=north east, ybar, symbolic x coords={1-gram, 2-gram, 3-gram, 4-gram}, ylabel = {\% of the duplicates}]
\addlegendentry{w/o DeconvDec}
\addplot 
coordinates{(1-gram, 18.7)(2-gram, 7.2)(3-gram, 4.3)(4-gram, 2.5)};
\addlegendentry{DeconvDec}
\addplot 
coordinates{(1-gram, 16.6)(2-gram, 4.4)(3-gram, 2.0)(4-gram, 1.0)};

\end{axis}
\end{tikzpicture}
\label{dup}
}%
\hspace{0.2in}
\subcaptionbox{BLEU scores on sentences of different lengths}[.45\textwidth]{
\begin{tikzpicture}
\selectcolormodel{cmyk}
\begin{axis}[legend pos=south west, xlabel={}, ylabel = {BLEU (\%)}, xmin=0, xmax=60, ymin=22, ymax=39, xtick=data, ytick={22,24,26,28,30,32,34,36,38,40}]
\addlegendentry{w/o DeconvDec}
\addplot 
coordinates{(0, 35.22)(10, 35.33)(20, 34.98)(30, 34.41)(40, 32.44)(50, 27.19)(60, 22.10)};

\addlegendentry{DeconvDec}
\addplot 
coordinates{
(0, 37.99)(10, 38.01)(20, 36.04)(30, 35.79)(40, 33.58)(50, 28.11)(60, 23.26)
};
\end{axis}
\end{tikzpicture}
\label{length}
}

\caption{\textbf{Percentage of the duplicates at sentence level and the BLEU scores on sentences of different lengths} Tested on the NIST 2003 dataset. The red bar and line indicate the performance of our model, and the blue bar and line indicate that of the attention-based SeqSeq model.}
\label{dupandlength}
\end{figure}

As our model generates translation with global information from the deconvolution-based decoder, it should learn to reduce repetition as it can learn to avoid generating same contents according to the conjecture by the deconvolution-based decoder about the target-side contexts. In order to test whether our model can mitigate the problem of repetition in translation, we test the repetition on the NIST 2003 dataset, following \citet{SeeEA2017}. The proportions of the duplicates of 1-gram, 2-gram, 3-gram and 4-gram in each sentence are calculated. Results on Figure \ref{dupandlength}(a) show that our model generates less repetitive translation. In particular, the proportion of duplicates of our model is less than half of that of the conventional Seq2Seq model. 

Moreover, to validate its robustness on different sentence-length levels, we test the BLEU scores on sentences of length no shorter than 10 to 60 of the NIST 2003 dataset. According to the results on Figure \ref{dupandlength}(b), though with the increase in length, the performance of our model is always stronger than the Seq2Seq model. However, with the increase of length, the advantage of our model becomes smaller. This is consistent with our hypothesis. Since the length of generation of the deconvolution-based decoder is assigned a particular value (30 words in Chinese-to-English translation) due to the limited computation resources, there is not enough global target-side information for translating long sentences (say, longer than 30 words). In our future work, we will delve into this problem and conduct further research to reduce computation cost.

\begin{figure*}[tb]
\begin{minipage}{.33\textwidth}
\centering
\includegraphics[height=4.5cm]{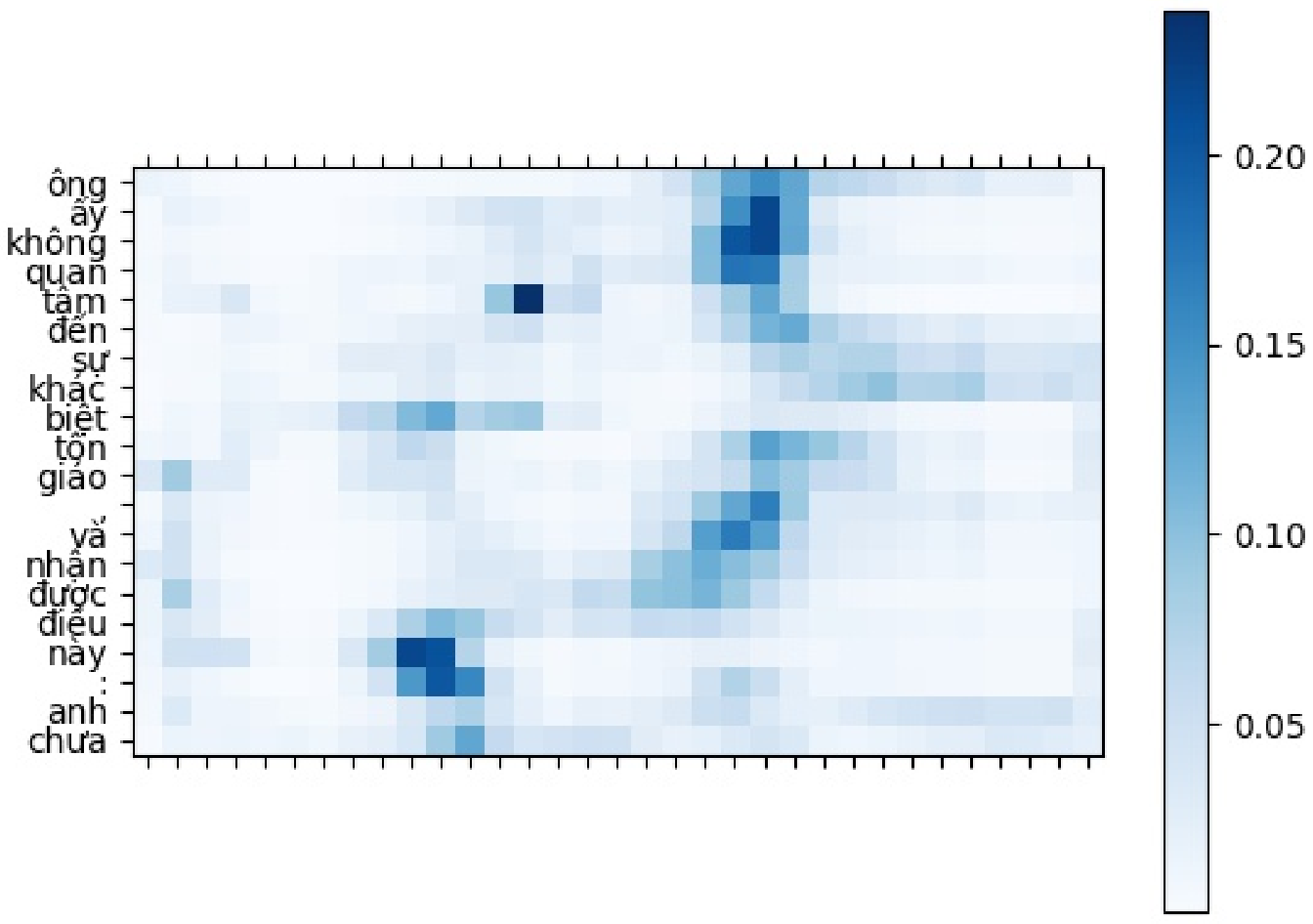}
\end{minipage}
\begin{minipage}{.33\textwidth}
\centering
\includegraphics[height=4.5cm]{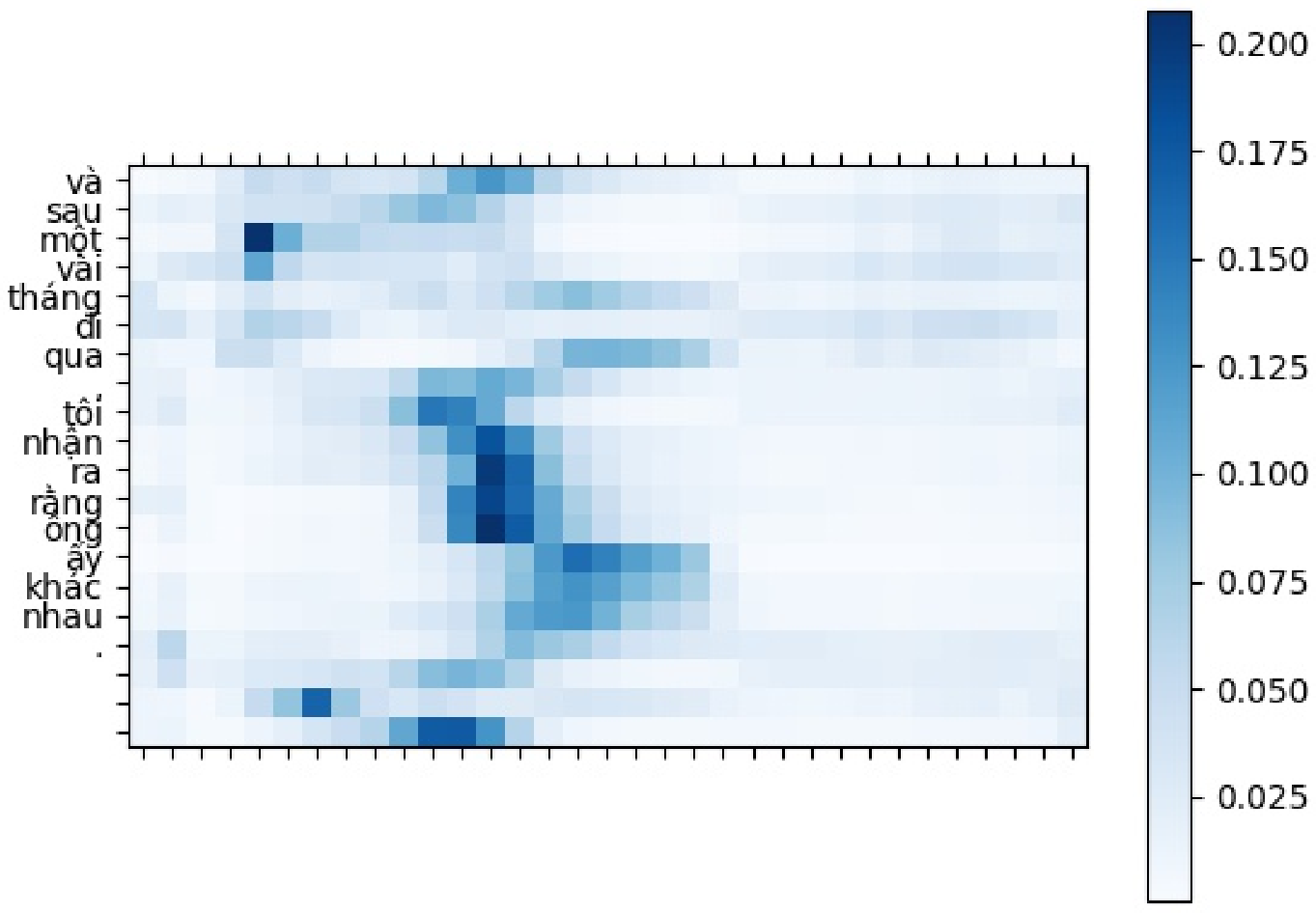}
\end{minipage}
\begin{minipage}{.33\textwidth}
\centering
\includegraphics[height=4.5cm]{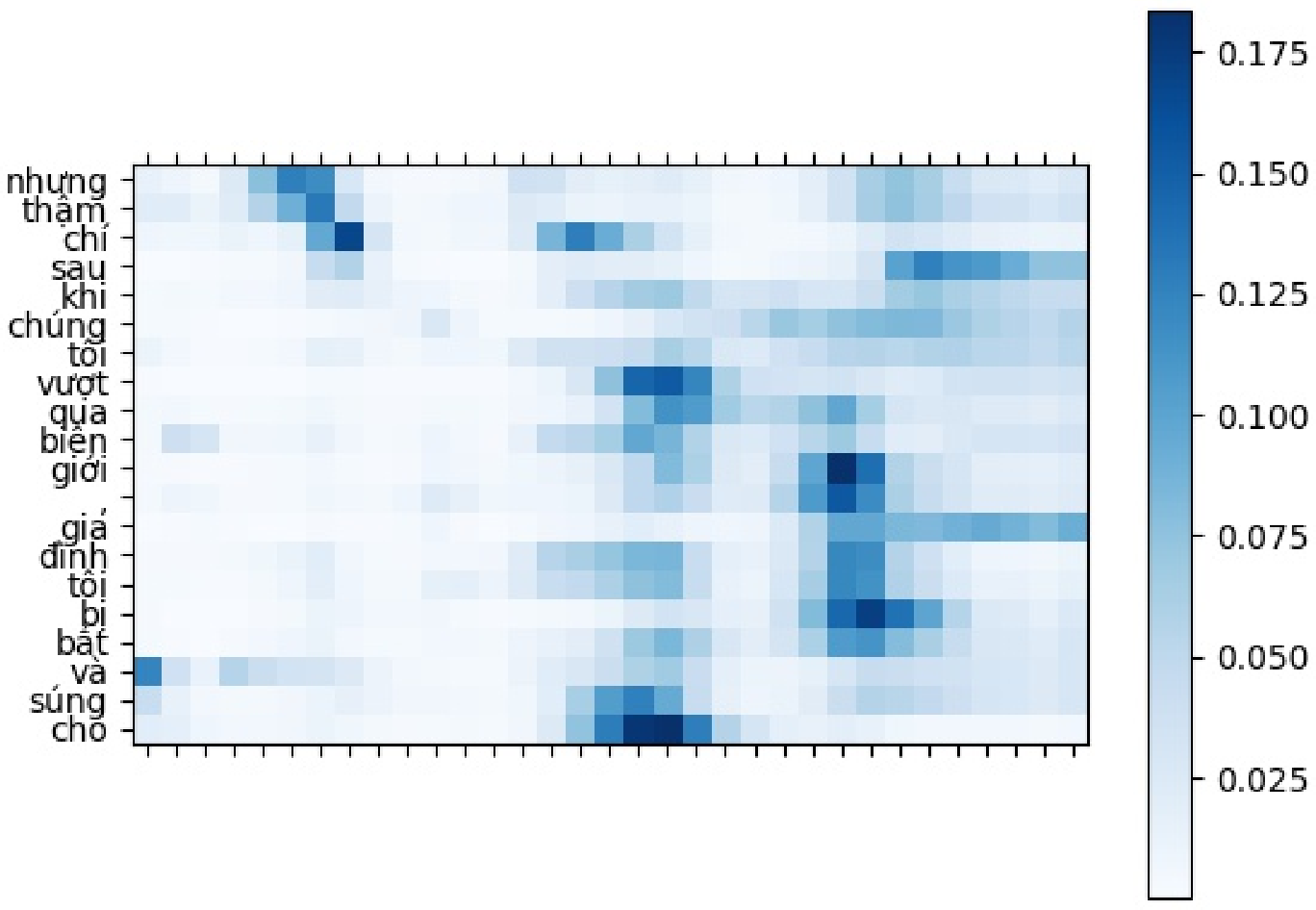}
\end{minipage}
\caption{\textbf{Attention heatmaps of the RNN-based decoder on the deconvolution-based decoder}\\ Words on the left refer to the translation of the RNN-based decoder. The heatmaps show that the RNN-based decoder can focus on certain parts of the outputs from the deconvolution-based decoder.}
\label{heat}
\end{figure*}

Figure \ref{heat} presents the attention heatmaps of the RNN-based decoder on the generated matrix of the deconvolution-based decoder in the English-to-Vietnamese translation. They reflect that the RNN-based decoder has diverse local focuses on the self-contained target-side contexts at different time steps. Contrary to the conventional attention on the source-side contexts which captures the corresponding annotations, it focuses on groups of the columns of the generated matrix from the deconvolution-based decoder. With the guidance of the information of global decoding, the model generates translation of higher accuracy and higher coherence. However, as the deconvolution-based decoder is not responsible for generating translation, it is hard to interpret what each column of the generated matrix represents. Moreover, as it does not capture alignment relationship as the conventional attention mechanism does, it is our future work to improve the attention on the outputs of the deconvolution-based decoder and explain the group focuses as shown in the heatmaps.

\subsection{Case Study}
\begin{table}[t]
\setlength{\tabcolsep}{3pt}
\centering
    \subcaptionbox{}{
    \begin{tabular}{p{15.5cm}}
    \hline
    \textbf{Text:} 基因 科学家 的 目标 是 , 提供 诊断 工具 以 发现 致病 的 缺陷 基因\\ 
    \hline
    \textbf{Gold:} the goal of geneticists is to provide diagnostic tools to identify defective genes that cause diseases\\
    \hline
    \textbf{Seq2Seq:} the objective of genetic scientists is to provide \colorbox[rgb]{0.99,0.86,0.86}{genes to detect genetic genetic genes}\\
    \hline
    \textbf{DeconvDec:} the objective of the gene scientist is to provide \colorbox{yellow}{diagnostic tools} to detect the \colorbox{yellow}{defective genes} \\
    \hline  
    \end{tabular}}
    \subcaptionbox{}{
    \begin{tabular}{p{15.5cm}}
    \hline
    \textbf{Text:} 叛军 暗杀 两位 菲 国 国会 议员 后, 菲律宾 总统 雅罗育 在 二零零一 年 中期 停止 与 共党 谈判 。\\ 
    \hline
    \textbf{Gold:} after the rebels assassinated two philippine legislators , philippine president arroyo ceased negotiations with the communist party in mid 2001 .\\
    \hline
    \textbf{Seq2Seq:} philippine president gloria arroyo stopped the two philippine parliament members in \colorbox[rgb]{0.99,0.86,0.86}{the mid - autumn festival} .\\
    \hline
    \textbf{DeconvDec:} philippine president gloria arroyo \colorbox{yellow}{stopped holding talks with the communist party} \colorbox{yellow}{after the rebels assassinated two philippine parliament members} .
\\
    \hline  
    \end{tabular}}
    \subcaptionbox{}{
    \begin{tabular}{p{15.5cm}}
    \hline
    \textbf{Text:} 中国 红十字会 已 在 24日 地震 发生 后 紧急 向 新疆 灾区 调拨 25万 元 人民币 , 又 于 25日 向 灾区 派出 救灾 工作组 。\\ 
    \hline
    \textbf{Gold:} china red cross has released 250 thousand renminbi for the xinjiang disaster area immediately after the earthquake on the 24th . a disaster relief team was dispatched to the area on the 25th .\\
    \hline
        \textbf{Seq2Seq:} \colorbox[rgb]{0.99,0.86,0.86}{the red cross society of china ( red cross ) , the china red cross society ( red cross )} , \colorbox[rgb]{0.99,0.86,0.86}{emergency} relief team sent an \colorbox[rgb]{0.99,0.86,0.86}{emergency} team to xinjiang for disaster relief in the disaster areas after the earthquake on 24 \colorbox[rgb]{0.99,0.86,0.86}{june} .\\
    \hline
    \textbf{DeconvDec:} the china red cross society has sent \colorbox{yellow}{250,000 yuan} to the disaster areas in xinjiang after the earthquake occurred on the 24 th, and sent a relief team to \colorbox[rgb]{0.99,0.86,0.86}{disaster disaster} areas on the 25 th.
\\
    \hline  
    \end{tabular}}
    \caption{Two examples of the translation of our model in comparison with the conventional attention-based Seq2Seq model on the NIST 2003 Chinese-to-English translation task. The errors in the translation are colored in red and the successful translation of some particular contents are colored in yellow (e.g., the contents that the model successfully translates but the other does not).}
    \label{example}
\end{table}
Table \ref{example} demonstrates three examples of the translation of our model in comparison with the Seq2Seq model and the golden translation. In Table \ref{example}(a), while the Seq2Seq model cannot recognize the objects of the main clause and the infinitive, causing inaccuracy and repetition, our model better captures the syntactic structure of the sentence and translates the main idea of the source text, though leaving the information ``that causes disease''. In Table \ref{example}(b), the source sentence is more complicated. With a temporal adverbial clause, its syntactic structure is more complex than simple sentence. The translation of the conventional Seq2Seq model cannot capture the syntactic information in the source text and regards the ``parliament members'' as the argument of the predicate ``talk''. Moreover, it is confused by the word ``中期'', meaning ``middle'', and translates ``mid - autumn festival''. In comparison, our model can recognize the adverbial clause and the main clause as well as their syntactic structures. In Table \ref{example}(c), it can be shown that when translating the long and relatively complex text, the baseline model makes a series of mistake of repetition. In contrast, the translation generated by our model though repeats the word ``disaster'', it is much more coherent and more semantically consistent with the source text as it successfully presents ``sent 250,000 yuan'' corresponding to the source text ``调拨25万元人民币'', while the baseline cannot translate the content. 

\section{Related Work}

In the following, we review the studies in NMT and the application of deconvolution in NLP.

\citet{Kalchbrenner,ChoEA2014,seq2seq} studied the application of the encoder-decoder framework on the machine translation task, which launched the development of NMT. Another significant innovation in this field is the attention mechanism, which builds connection between the translated contents and the source text \citep{attention,stanfordattention}. To improve the quality of NMT, researchers have focused on improving the attention mechanism. \citet{coverage} and \citet{micover} modeled coverage in the NMT, \citet{interactive} and \citet{multichannel} incorporated the external memory to the attention, and \citet{target_attention} as well as \citet{aca} utilized the information from the previous generation by target-side attention and memory for attention history respectively. For more target-side information, \citet{BOW_Ma} incorporated bag of words as target. A breakthrough of NMT in recent years is that \citet{googleattention} invented a model only with the attention mechanism that reached the state-of-the-art performance.

Although many researches in NLP focused on the application of RNN, CNN is also an important type of network for the study of language \citep{kimconv, kalconv, zhangconv, global_encoding}. Also, its application in NMT has been successful \citep{fairseq}. Recently, deconvolution was applied to modeling text \citep{deconvolution,deconvseq}, which is able to construct a representation of high quality with the self-contained information.

\section{Conclusion and Future Work}
\label{conclusion}

In this paper, we propose a new model with the global decoding mechanism. With our deconvolution-based decoder, which provides global information of the target-side contexts, the model can effectively exploit the information for the inference of syntactic structure and semantic meaning in the translation. Experimental results on the Chinese-to-English translation and English-to-Vietnamese translation demonstrate that our model outperforms the baseline models, and the analysis shows that our model generates less repetitive translation and demonstrates higher robustness to the sentences of different lengths. Furthermore, the case study shows that the translation of our model better observes the syntactic and semantic requirements for the translation and generates coherent and accurate translation with fewer irrelevant contents.

In the future, we will further develop analysis of the mechanism of deconvolution in NMT and try to figure out its generalized patterns for the construction of the target-side contexts.

\section*{Acknowledgements}
This work was supported in part by National Natural Science Foundation of China (No. 61673028) and the National Thousand Young Talents Program. Xu Sun is the corresponding author of this paper.

\bibliographystyle{coling}
\bibliography{coling2018}








\end{CJK}

\end{document}